\documentclass[preprint,12pt]{elsarticle}

\usepackage{amssymb}
\usepackage{amsmath}
\usepackage[ruled,vlined,linesnumbered]{algorithm2e}
\usepackage{booktabs}
\usepackage{pifont}
\usepackage{hyperref}
\hypersetup{hypertex=true,
colorlinks=true,
linkcolor=blue,
anchorcolor=blue,
citecolor=blue}

\usepackage{xcolor}
\usepackage[table]{xcolor}
\usepackage{colortbl}
\usepackage{caption}
\newcommand{\revisee}[1]{#1}
\journal{Nuclear Physics B}

\begin{document}

\begin{frontmatter}



\title{Instruction-as-State: Environment-Guided and State-Conditioned Semantic Understanding for Embodied Navigation}


\author[a,c]{Zhen Liu\fnref{fn1}} %% Author name
\author[a,c]{Yuhan Liu} %% Author name
\author[a,c]{Jinjun Wang} %% Author name
\author[a,c]{Jianyi Liu\fnref{fn2}}
\author[d]{Wei Song} %% Author name
\author[e]{Jingwen Fu\fnref{fn3}} %% Author name

\affiliation[a]{organization={National Key Laboratory of Human-Machine Hybrid Augmented Intelligence},
            country={China}}

\affiliation[c]{organization={The Institute of Artificial Intelligence and Robotics},
            addressline={Xi'an Jiaotong University, Xian Ning West Road No. 28}, 
            city={Xi'an},
            postcode={710049}, 
            state={Shaanxi},
            country={China}}

\fntext[fn1]{E-mail: zhenliu9773@gmail.com.}

\fntext[fn2]{Corresponding author. E-mail: jyliu@xjtu.edu.cn.}

\fntext[fn3]{Project leader. E-mail: fujingwen@bza.edu.cn.}

\affiliation[d]{organization={School of Information Science and Technology},
            addressline={North China University of Technology}, 
            postcode={100144}, 
            state={Beijing},
            country={China}}

\affiliation[e]{organization={Zhongguancun Academy},
            addressline={Zhongguancun},
            city={Beijing},
            postcode={100190},
            state={Beijing},
            country={China}}

%% Abstract
\begin{abstract}

\revisee{Vision-and-Language Navigation requires agents to follow natural-language instructions in visually changing environments.}
A central challenge is the dynamic entanglement between language and observations: the meaning of instruction shifts as the agent's field of view and spatial context evolve.
\revisee{However, many existing models encode the instruction as a static global representation, limiting their ability to adapt instruction meaning to the current visual context.}
\revisee{We therefore model instruction understanding as an Instruction-as-State variable: a decision-relevant, token-level instruction state that evolves step by step conditioned on the agent's perceptual state, where the perceptual state denotes the observation-grounded navigation context at each step.}
To realize this principle, we introduce State-Entangled Environment-Guided Instruction Understanding (S-EGIU), a coarse-to-fine framework for state-conditioned segment activation and token-level semantic refinement.
At the coarse level, S-EGIU activates the instruction segment whose semantics align with the current observation.
At the fine level, it refines the activated segment through observation-guided token grounding and contextual modeling, sharpening its internal semantics under the current observation.
\revisee{Together, these stages maintain an instruction state that is continuously updated according to the agent's perceptual state during navigation.}
S-EGIU delivers strong performance on several key metrics, including a +2.68\% SPL gain on REVERIE Test Unseen, and demonstrates consistent efficiency gains across multiple VLN benchmarks, underscoring the value of dynamic instruction--perception entanglement.
\end{abstract}

%% Keywords

\end{frontmatter}




\section{Introduction}

VLN requires agents to interpret natural-language instructions while operating in continuously changing visual environments~\cite{Wu2021, Li2023}.
\revisee{To execute an instruction, an agent must ground landmarks, spatial relations, and actions in its real-time visual observations, meaning that the usefulness of each instruction cue depends on the agent's perceptual state, which refers to the observation-grounded navigation context at the current step~\cite{anderson2018vision, pashevich2021episodic}.}
A central challenge arises from the dynamic entanglement between instruction and perception: \revisee{instruction semantics are not interpreted independently of the visual context, but are continuously shaped by what the agent currently sees. An instruction cue becomes meaningful only when its associated scene elements enter view, and its semantic role may shift as visibility and spatial context evolve.}
However, most existing VLN models~\cite{He2023, Schumann2023} encode the entire instruction as a static global representation, preventing the model from updating instruction relevance as the agent moves through the environment.
\revisee{This disconnect from the evolving perceptual state hinders the agent's ability to determine which instruction components are currently applicable.}

Although prior VLN studies have explored joint text–vision embeddings~\cite{ma2019self}, 
semantic decomposition~\cite{Qi2020}, and data augmentation through speaker models~\cite{Fried2018}, 
these approaches all rely on a static, full-instruction representation that remains fixed throughout the navigation trajectory. 
\revisee{This assumption contradicts the nature of embodied navigation, where the relevance of each instruction segment depends on the agent's current observation: a landmark becomes meaningful only when it enters view, directional cues vary with orientation, and spatial relations must be reinterpreted as the environment changes.}

Recent efforts have attempted to mitigate this limitation by incorporating sub-instruction structures. 
Sub-Instruction Aware VLN~\cite{Hong2020} decomposes the full instruction into smaller textual segments and models their sequential order, MLANet~\cite{He2023} introduces multi-level attention 
to highlight phrase-level cues, and the Sub-Instruction and Local Map Relationship Enhanced Model~\cite{Zhang2023} integrates local map priors to strengthen spatial grounding. 
While these approaches introduce useful structural decomposition, their segmentation and activation mechanisms remain text-driven or map-driven \revisee{rather than conditioned on the agent's perceptual state}. 
As a result, sub-instructions may be activated prematurely (before their referents enter view) or may linger after the visual evidence disappears, limiting their ability to track which instruction fragment is truly relevant under the current viewpoint.
Because static or predefined sub-instruction structures cannot adjust to the agent’s evolving observations, existing VLN models fail to reveal which parts of the instruction are currently applicable or when new segments should become active. 
\revisee{This mismatch motivates our Instruction-as-State perspective.
Instead of treating the instruction as a fixed sentence embedding, we model instruction understanding as a dynamic latent state for decision making whose content is updated at each step conditioned on the agent's perceptual state (i.e., the observation-grounded navigation context).
We use the term ``state'' to emphasize that this instruction representation is trajectory-evolving and decision-relevant, rather than a static textual encoding.}

\begin{figure*}[h]
    \centering
    \includegraphics[width=\textwidth]{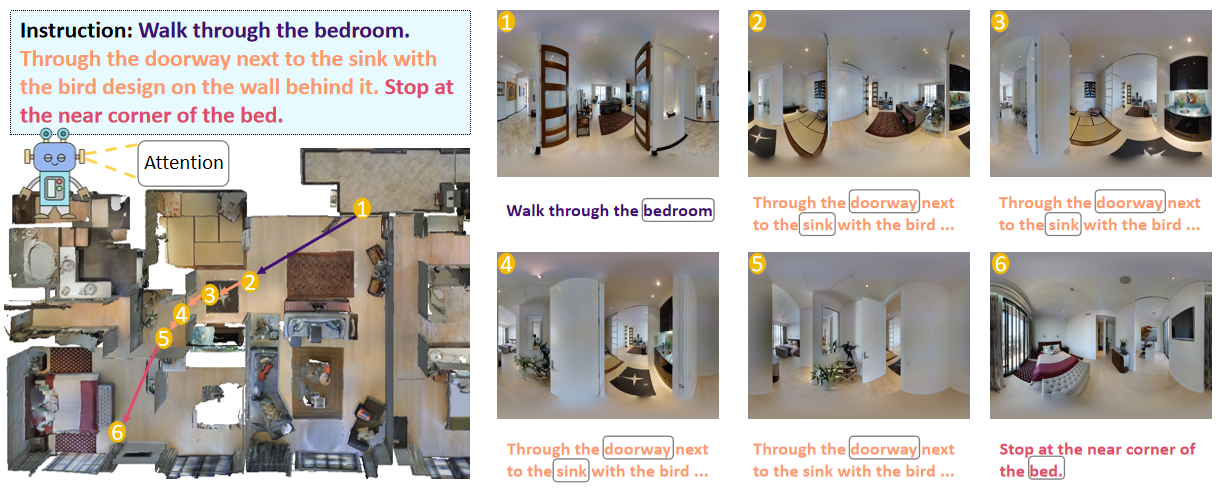}%
    \label{fig_instruction_following}
    \caption{Example of dynamic instruction semantics in VLN. 
At each viewpoint, the agent selects a perceptually relevant sub-instruction (color-coded) and refines its meaning by emphasizing key tokens, illustrating the Instruction-as-State behavior modeled by S-EGIU.}
    \label{fig:instruction_following}
\end{figure*}

VLN instructions typically contain multiple navigation steps and targets \revisee{whose relevance depends on the agent’s current observation.}
For example, the instruction “\textit{Walk through the bedroom. Through the doorway next to the sink ...}” encodes a sequence of semantics whose relevance depends on the observation-grounded context that defines the perceptual state at different steps (see Fig.~\ref{fig:instruction_following}).
As the agent approaches the bedroom, the sub-instruction “\textit{Walk through the bedroom}” \revisee{becomes the currently active instruction segment, while downstream clauses remain inactive until their associated landmarks enter view.}
Within the active sub-instruction, token-level importance, such as the relevance of “\textit{bedroom}”, also shifts as new visual evidence appears.
This sub-instruction-level activation and token-level modulation illustrate the dynamic entanglement between instruction semantics and the agent’s observation-grounded context, demonstrating that instruction meaning cannot remain globally fixed.

\revisee{To address this challenge, we propose the State-Entangled Environment-Guided Instruction Understanding (S-EGIU) framework, which enables the dynamic interaction between instruction state and the agent’s evolving perceptual state.}
Instead of treating the instruction as a fixed, globally relevant description, S-EGIU uses real-time perceptual input to identify and update the instruction components that are meaningful at each point along the navigation trajectory. \revisee{Under this formulation, the instruction is treated as a dynamic latent state variable for decision making rather than a static sentence-level representation.}

S-EGIU operates in two hierarchical stages: Coarse-Grained Instruction Processing (CGIP) and Fine-Grained Instruction Processing (FGIP).
CGIP segments the instruction into structured sub-instructions and activates the one \revisee{that is most relevant under the agent's perceptual state, while suppressing content whose relevance has not yet emerged.}
FGIP then performs perception-guided semantic refinement on the active sub-instruction, adjusting token-level importance so that spatial references, entities, and directional semantics reflect what the agent currently observes.
\revisee{Through this coarse-to-fine refinement conditioned on the agent’s perceptual state,} S-EGIU enables instruction state to evolve naturally with perception, supporting reliable guidance in both seen and unseen environments.

We evaluate S-EGIU across three widely used VLN benchmarks—R2R, SOON, and REVERIE—which collectively span short-horizon navigation, long compositional instructions, and object-centric grounding.
Across datasets, S-EGIU consistently improves trajectory efficiency while maintaining competitive navigation success against strong baselines.
Notably, it achieves strong performance, including a +2.68\% SPL gain on REVERIE Test Unseen, \revisee{demonstrating the advantage of dynamically updating the instruction state under the perceptual state.}

The key contributions of this work are summarized as follows:
\begin{itemize}
\item
We propose S-EGIU, a framework that models instruction understanding \revisee{as a dynamically evolving process conditioned on the agent's perceptual state, overcoming the limitation of static instruction encodings in VLN.}
\item
We introduce a coarse-to-fine instruction modeling architecture that operationalizes the Instruction-as-State perspective:
\revisee{CGIP activates the sub-instruction that is most relevant under the perceptual state,} while FGIP performs perception-guided refinement of the active sub-instruction.
\item
We demonstrate the effectiveness of S-EGIU on R2R, SOON, and REVERIE, achieving consistent efficiency gains while maintaining competitive navigation success across these benchmarks, including a +2.68\% SPL gain on REVERIE Test Unseen.
\end{itemize}
\section{Related Work}

\subsection{Vision-and-Language Navigation}

Vision-and-Language Navigation (VLN) aims to enable embodied agents to follow natural-language instructions in visually complex indoor environments. Early VLN systems relied on handcrafted visual features and rule-based navigation heuristics, which offered limited generalization to unseen environments. The Room-to-Room (R2R) benchmark~\cite{anderson2018vision} shifted VLN toward end-to-end learning, enabling neural agents to map natural-language instructions to navigation actions at scale. Subsequent research has advanced VLN from multiple perspectives, including cross-modal grounding, long-horizon reasoning, global planning, and data augmentation. Attention- and memory-based models~\cite{wang2021ssm, zhang2022exor} enhanced grounding by integrating textual semantics with panoramic visual observations. Transformer-based agents and map-centered architectures have further advanced long-horizon reasoning and global planning in VLN~\cite{chen2022think, wang2021ssm,zhang2025causal,wang2025instruction}. Data augmentation approaches such as Speaker Follower~\cite{Fried2018} further improved grounding by generating synthetic instructions. 

\revisee{Recent VLN research has also begun to explore large language models (LLMs) and large vision-language models (LVLMs) for embodied navigation. Representative works such as NavGPT~\cite{zhou2024navgpt}, NaVid~\cite{zhang2024navid}, and NavGPT-2~\cite{zhou2024navgpt2} incorporate foundation-model-based reasoning, multimodal grounding, or large-scale visual-language pretraining to enhance navigation decision making. These methods highlight the growing importance of explicit reasoning and general-purpose multimodal representations in VLN, as also summarized in recent survey literature~\cite{zhang2024vlnsurvey}.}

\subsection{Instruction Understanding in VLN}

Instruction modeling plays a central role in VLN, and prior work has explored a range of language-encoding strategies aimed at improving cross-modal grounding. Early systems used recurrent networks for sequential instruction encoding~\cite{anderson2018vision}, followed by attention-based models that highlight potentially relevant phrases within long instructions~\cite{hao2020prevalent}. Transformer-based encoders~\cite{guhur2021airbert} improve global reasoning over instruction tokens, while graph-based models~\cite{chen2021instruction} capture multi-scale semantic structure to enhance cross-modal grounding. Memory-driven architectures such as Episodic Transformer~\cite{pashevich2021episodic} and Think-Global, Act-Local~\cite{chen2022think} further enhance instruction grounding by incorporating episodic or persistent contextual memory. 

\revisee{Despite these advances in language encoding and cross-modal fusion, many existing VLN methods still rely on globally encoded instruction representations, without explicitly modeling how instruction relevance evolves with the agent's perceptual state. Even methods with progress estimation or dynamic attention~\cite{wang2021ssm,hao2020prevalent} mainly adjust token importance or cross-modal alignment, rather than explicitly updating instruction semantics as new visual evidence appears during navigation. As a result, they provide limited support for determining which sub-instructions are currently most applicable and how their semantic relevance should be updated step by step. Motivated by this limitation, our work explicitly models instruction understanding under the perceptual state and updates the semantics of relevant sub-instructions accordingly during navigation.}

\section{Method}

In this section, we first formalize the VLN task and then present our State-Entangled Environment-Guided Instruction Understanding (S-EGIU) framework.
\revisee{S-EGIU is built upon the instruction-as-state principle, which treats the instruction not as a fixed text embedding but as an evolving instruction state that is updated in a step-wise manner conditioned on the agent's perceptual evidence along the trajectory.}
To operationalize this principle, S-EGIU performs a coarse-to-fine update at every navigation step through two modules:
\revisee{Coarse-Grained Instruction Processing (CGIP), which estimates a clause-relevance distribution under the current visual evidence and activates the most relevant sub-instruction through top-1 routing, and Fine-Grained Instruction Processing (FGIP), which refines the token-level semantics of this activated sub-instruction using current visual evidence.}
Together, these components produce a dynamically evolving instruction state that supports perception-aligned navigation decisions throughout the trajectory.

\subsection{VLN Task Definition}

VLN requires an embodied agent to navigate in a photorealistic 3D environment by following a natural-language instruction. Let the instruction be a token sequence $I = \{w_1,\dots,w_L\}$. \revisee{At each time step $t$, the agent observes a 36-view panoramic observation and encodes it as a visual feature $V_t\in\mathbb{R}^{N\times d}$ (perceptual state), where $N$ denotes the number of visual tokens and $d$ is the hidden dimension, and selects an action $a_t$ from the set of navigable viewpoints.} The objective is to execute the instruction by producing a trajectory $\tau = (a_1,\dots,a_T)$ that reaches the target location.

\subsection{State-Entangled Environment-Guided Instruction Understanding}
\label{sec:S-EGIU}
\begin{figure*}[t]
    \centering
    \includegraphics[width=\textwidth]{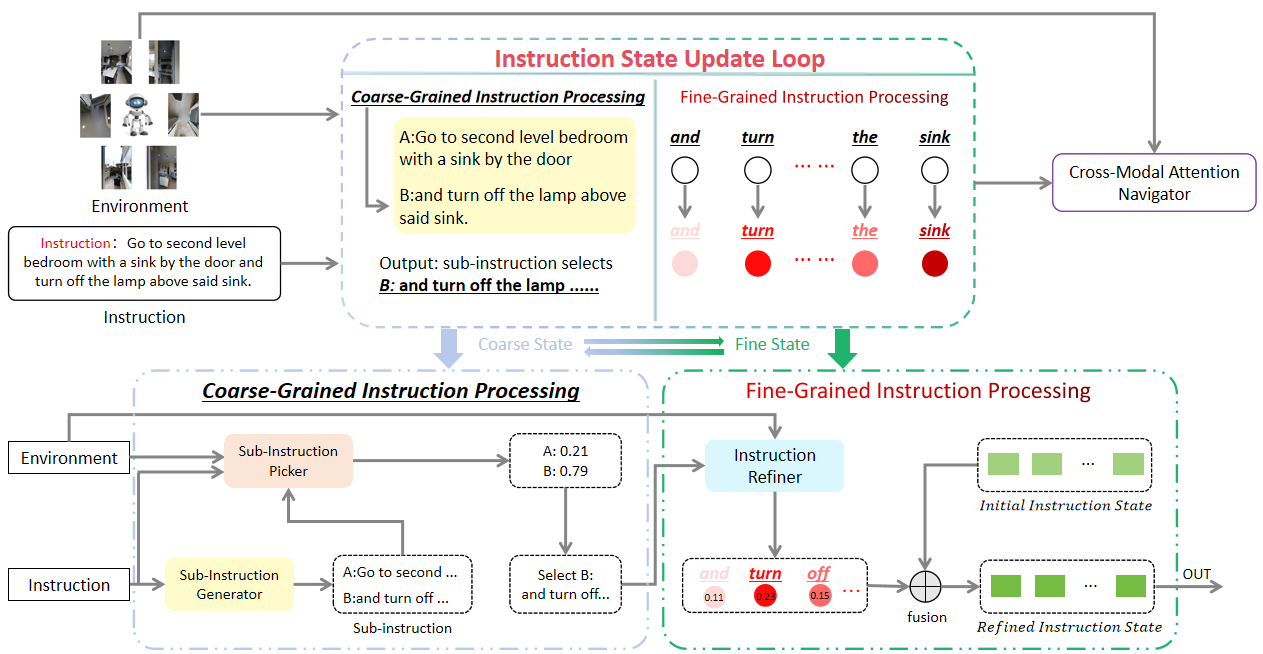}%
\caption{Overview of the S-EGIU framework. CGIP estimates a clause-relevance distribution under the current visual context and activates the most strongly grounded sub-instruction through top-1 routing. FGIP then applies perception-guided token refinement to update the instruction state, which is passed to the navigation policy.}
    \label{fig:new_framework}
\end{figure*}
\revisee{Building on the Instruction-as-State formulation, S-EGIU updates the instruction state at each navigation step through a two-stage coarse-to-fine procedure (Fig.~\ref{fig:new_framework}).
At each navigation step $t$, the model maintains a token-level instruction state
$S_{t-1} \in \mathbb{R}^{L \times d}$ and the current panoramic observation $V_t$,
where $L$ is the instruction length and $d$ is the hidden dimension.
The initial state $S_0$ is derived from the full-instruction encoder output.
CGIP first performs sub-instruction-level reasoning on the full instruction to estimate a clause-relevance distribution under the current observation $V_t$, and activates the most relevant segment through top-1 routing.
FGIP then refines the activated segment through perception-guided token grounding and contextual encoding, and fuses the refined representation back into the full instruction-state space to obtain the updated instruction state $S_t$.}

For example, in the instruction
``\textit{Go to the second-level bedroom with a sink by the door and turn off
the lamp above the sink}'',
\revisee{only the latter sub-instruction becomes perceptually grounded when the sink and lamp enter view.}
\revisee{CGIP identifies this sub-instruction, and FGIP refines its token representations conditioned on the current observation $V_t$, producing an instruction state aligned with the agent’s perceptual state.}
\revisee{This example also illustrates why panoramic observation is used as the perceptual state: relevant landmarks and directional cues may appear across different viewing directions. Potentially irrelevant panoramic context is mitigated by first activating the grounded sub-instruction in CGIP and then refining only its associated token states in FGIP.}
\revisee{Algorithm~\ref{alg:S-EGIU} summarizes the update process and, for readability, presents the inference-time top-1 routing; during training, the same activation is implemented via a straight-through relaxation over the clause-relevance distribution $\alpha_t$.}

\begin{algorithm}[t]
\caption{Coarse-to-Fine Instruction-State Update in S-EGIU}
\label{alg:S-EGIU}

\revisee{\KwIn{Instruction $I$, observation $V_t$, instruction states $S_0$ and $S_{t-1}$}}

\KwOut{Updated instruction state $S_t$}

\BlankLine
\textbf{Stage 1: CGIP}\

\textit{Sub-Instruction Generator:}\

\revisee{$\mathcal{C} \leftarrow \mathrm{Segment}(I)$}

\revisee{$\hat{\mathcal{C}} \leftarrow \mathrm{RefineSeg}(\mathcal{C})$}

\textit{Sub-Instruction Picker:}\

\revisee{$r \leftarrow \mathrm{Score}(\mathrm{CrossAttn}(S_0, V_t))$}

\revisee{$\phi_{t,k} \leftarrow \sum_{i \in T_k} w_t(i)\, r_i$}

\revisee{$\alpha_t \leftarrow \mathrm{softmax}(\phi_t),$}

\revisee{$k^\ast \leftarrow \arg\max_k \alpha_{t,k}$}

\BlankLine
\textbf{Stage 2: FGIP}\

\textit{Instruction Refiner:}\

\revisee{$\tilde{T} \leftarrow \mathrm{CrossAttn}(S_{t-1}[T_{k^\ast}], V_t)$}

\revisee{$\hat{R}_t \leftarrow \mathrm{Trans}(\tilde{T})$}

\revisee{$R_t \leftarrow \mathrm{Scatter}(\hat{R}_t, T_{k^\ast})$}

\textit{Fusion:}\

\revisee{$S_t \leftarrow S_{t-1} + g \odot (R_t - S_{t-1})$}

\Return{$S_t$}

\end{algorithm}

\subsection{Instruction-as-State Formulation}

\revisee{We formulate instruction understanding as the step-wise update of an evolving instruction state conditioned on the agent’s perceptual state along the trajectory.}
\revisee{Let $S_0 \in \mathbb{R}^{L \times d}$ denote the initial instruction state derived from the full instruction $I$.
At each navigation step $t$, the agent observes a panoramic visual feature $V_t$ and updates
the previous instruction state $S_{t-1}$ to a new state $S_t$ via
\begin{equation}
\label{eq:ias}
S_t = f(S_{t-1},\, S_0,\, I,\, V_t).
\end{equation}}

\revisee{The update function consists of two operations:
(i) identifying which sub-instruction is currently relevant under the observation, and
(ii) refining its token-level semantics based on perceptual state.
To describe these operations, we denote by $\mathrm{CrossAttn}(X, Y)$ a cross-attention operation
in which $X$ provides the queries and $Y$ provides the keys and values.
Accordingly, we decompose the step-wise update into a coarse-to-fine structure:
\begin{equation}
T_{k^\ast} = f_{\text{CGIP}}(S_0,\, I,\, V_t),
\end{equation}
\begin{equation}
S_t = f_{\text{FGIP}}(S_{t-1},\, T_{k^\ast},\, V_t).
\end{equation}
Here, $T_{k^\ast}$ denotes the token index set of the sub-instruction activated by CGIP.
CGIP operates on the initial instruction state $S_0$ as a stable semantic anchor for coarse clause-level selection.
We intentionally avoid using the evolving state $S_{t-1}$ in this stage, because early grounding errors may accumulate over time and progressively bias subsequent coarse sub-instruction activation.
In contrast, FGIP operates on the previous-step state $S_{t-1}$ to perform observation-conditioned fine-grained refinement within the activated segment and fuse it back into the full instruction state.
This coarse-stable and fine-adaptive decomposition improves robustness while preserving dynamic instruction-state evolution.}

This formulation establishes the theoretical foundation of S-EGIU: instruction
semantics are represented as a state that evolves with the agent’s viewpoint,
rather than as a fixed text embedding, enabling dynamic entanglement between
instruction interpretation and real-time perception.

\subsection{Coarse-Grained Instruction Processing}

\revisee{VLN instructions are often compositional, containing multiple semantic clauses whose relevance depends on the agent’s perceptual state.}
\revisee{CGIP serves as the first stage of the Instruction-as-State hierarchy and produces a coarse structural prior by estimating a clause-relevance distribution under the current visual observation and activating the most strongly grounded sub-instruction at step~$t$.}
By filtering out perceptually invalid or inactive clauses, CGIP reduces the semantic hypothesis space and establishes a grounded structural scaffold for the fine-grained refinement executed by FGIP.

\paragraph{Sub-Instruction Generator}
Given the full instruction $I$, CGIP first constructs a structured set of
candidate semantic units through a two-stage segmentation procedure. The
goal is to recover clause-level navigation steps that align with the
instruction’s underlying temporal and spatial structure.

\textbf{Rule-based coarse segmentation.}
We begin with a lightweight rule-based clause detector,
\revisee{\begin{equation}
\mathcal{C} = \{C_k\}_{k=1}^m = \mathrm{Segment}(I),
\end{equation}}
\revisee{where $\{C_k\}_{k=1}^m$ denotes the initial set of coarse instruction clauses produced by the rule-based segmentation module.}
\revisee{The segmentation module identifies candidate boundaries associated with common navigational
conjunctions (e.g., ``and'', ``then'', ``after'') and structural markers such as
punctuation. This step captures high-level semantic transitions typically
found in VLN instructions, yielding an initial set of coarse clauses. However,
real-world instructions exhibit diverse syntactic patterns, and conjunctions
may appear inside descriptive phrases (e.g., ``the room with a table and a
chair''), causing over-segmentation or missed boundaries.}

\textbf{Learnable boundary refinement.}
To obtain segments that more faithfully correspond to true navigation
subgoals, we introduce a learnable refinement module:
\begin{equation}
\hat{\mathcal{C}} = \{\hat{C}_k\}_{k=1}^{\hat{m}} = \mathrm{RefineSeg}(\mathcal{C}).
\end{equation}

\revisee{Let the instruction $I$ contain $L$ tokens. The coarse candidate set
$\mathcal{C}$ produced by $\mathrm{Segment}$ induces an initial boundary set
$\mathcal{B}^{(0)} \subseteq \{1,\dots,L-1\}$, where each boundary position
$i$ denotes the gap between the $i$-th and $(i+1)$-th tokens. Based on this
coarse segmentation, $\mathrm{RefineSeg}$ re-evaluates each inter-token
candidate boundary position and predicts a refined boundary confidence as}
\begin{equation}
\revisee{\hat{b}_i = \sigma\!\big(\mathrm{MLP}([\,h_i;\,h_{i+1};\,p_i;\,\psi_i\,])\big),
\quad i=1,\dots,L-1,}
\end{equation}
\begin{equation}
\revisee{\hat{\mathcal{B}}=\{\, i \mid \hat{b}_i > \delta_b,\; i=1,\dots,L-1 \,\},}
\end{equation}
\begin{equation}
\revisee{\hat{\mathcal{C}}=\mathrm{Split}(I,\hat{\mathcal{B}}),\qquad
\hat{m}=|\hat{\mathcal{B}}|+1,}
\end{equation}
\revisee{where $h_i$ and $h_{i+1}$ are the contextualized representations of
the two neighboring tokens from the instruction encoder, $p_i=\mathbf{1}[i\in
\mathcal{B}^{(0)}]$ is the coarse boundary prior induced by $\mathcal{C}$,
$\mathbf{1}[\cdot]$ denotes the indicator function, and $\psi_i$ is a local
semantic coherence cue derived from neighboring token representations.
Here, $\delta_b$ is a fixed boundary threshold, $\hat{\mathcal{B}}$ is the
set of predicted refined boundary positions, and $\mathrm{Split}(\cdot)$
partitions the instruction token sequence according to these boundary
positions to form the refined candidate set $\hat{\mathcal{C}}$, where
$\hat{m}$ is the number of refined sub-instructions.}

By jointly considering these signals, $\mathrm{RefineSeg}$ suppresses spurious
rule-based boundaries and recovers missing boundaries when the semantics
indicate a genuine shift in navigational intent.
\revisee{The module does not require explicit boundary annotations. Instead,
the boundary scorer is learned from the same training data as the navigation
model, with the rule-based boundaries providing coarse structural priors and
the downstream navigation objective encouraging task-relevant boundary
refinement.}
\revisee{This yields a refined and semantically coherent candidate set
$\hat{\mathcal{C}}$, which is passed to the subsequent sub-instruction
selection stage in Algorithm~\ref{alg:S-EGIU}.}

\paragraph{Sub-Instruction Picker}
To determine which refined sub-instruction is currently grounded in the
visual observation, CGIP evaluates the perceptual relevance of each
candidate. We first compute token--region alignment via multi-head
cross-attention between the initial instruction state and the panoramic
visual feature:
\revisee{\begin{equation}
r = \mathrm{Score}\!\left(\mathrm{CrossAttn}(S_0, V_t)\right),
\end{equation}}
\revisee{where $S_0$ provides the queries and $V_t$ provides the keys and values. Let $U_t=\mathrm{CrossAttn}(S_0,V_t)$ denote the resulting cross-attended token representations. We then instantiate $\mathrm{Score}(\cdot)$ as a lightweight token-wise predictor applied to $U_t$:
\begin{equation}
\label{eq:score}
r=\sigma(U_t W_r + b_r),
\end{equation}
where $W_r\in\mathbb{R}^{d\times 1}$ and $b_r\in\mathbb{R}$ are learnable parameters and $\sigma(\cdot)$ is the sigmoid function. This yields a token-wise visual relevance vector $r\in[0,1]^{L}$, where the scalar $r_i$ measures how strongly token $i$ is grounded under the current observation $V_t$.}

\revisee{For each refined sub-instruction $\hat{C}_k$, let $T_k \subseteq \{1,\dots,L\}$ denote the set of token positions belonging to $\hat{C}_k$. We aggregate the token-wise relevance scores $\{r_i \mid i \in T_k\}$ with step-dependent weighting to obtain a state-conditioned groundedness score:}
\begin{equation}
\revisee{\phi_{t,k} = \sum_{i \in T_k} w_t(i)\, r_i,}
\end{equation}
\revisee{where $\phi_{t,k}$ measures how strongly the semantics of $\hat{C}_k$ are supported by the current observation $V_t$.}
\revisee{Here, $w_t(i)$ is a {clause-wise normalized token-importance weight} computed at step $t$, which modulates each token's contribution when aggregating token groundedness into the segment-level score $\phi_{t,k}$.
Intuitively, even within a candidate clause, only a subset of tokens is informative under the current observation (e.g., landmarks or spatial relations), while others contribute less.
In implementation, we derive $w_t(i)$ from the cross-attended token representations $U_t$ introduced above. Specifically, we compute an unnormalized token importance score $\tilde{w}_t(i)=\exp(\mathrm{MLP}_w(U_t[i]))$ and normalize it within each candidate sub-instruction:
\begin{equation}
\label{eq:wt}
w_t(i)=\frac{\tilde{w}_t(i)}{\sum_{j\in T_k}\tilde{w}_t(j)} \ \ (i\in T_k),
\end{equation}
so that $w_t(i)\in(0,1)$ and $\sum_{i\in T_k}w_t(i)=1$. 
This within-clause normalization stabilizes the groundedness aggregation and emphasizes the most informative tokens when computing $\phi_{t,k}$, and it is used only for within-step clause aggregation rather than any across-step temporal smoothing.}

\revisee{The clause-relevance distribution is computed as}
\begin{equation}
\revisee{\alpha_t = \mathrm{softmax}(\phi_t),}
\end{equation}
\revisee{where $\phi_t = [\phi_{t,1}, \dots, \phi_{t,\hat m}] \in \mathbb{R}^{\hat m}$ denotes the clause-score vector, and $\alpha_t \in \Delta^{\hat m-1}$ is the resulting differentiable clause-relevance distribution.
During training, we adopt a straight-through top-1 routing strategy: the forward pass uses}
\begin{equation}
\revisee{k^\ast = \arg\max_k \alpha_{t,k},}
\end{equation}
\revisee{to preserve discrete active-clause reasoning in FGIP, while gradients are back-propagated through $\alpha_{t,k}$.
During inference, we use the same top-1 routing without the straight-through gradient estimator.}

\revisee{The activated $\hat{C}_{k^\ast}$ defines the coarse structural prior by
retaining only perceptually supported content while suppressing inactive or
irrelevant clauses. Equivalently, CGIP outputs the token index set $T_{k^\ast}$
and its corresponding token mask, which together serve as the structural prior
for the fine-grained semantic refinement in FGIP.}

\subsection{Fine-Grained Instruction Processing}

While CGIP identifies which sub-instruction is perceptually grounded at
step~$t$, the tokens within that sub-instruction do not contribute equally
to navigation. 
\revisee{Landmarks, directional terms, and spatial relations become more informative once their visual referents are observed, whereas less grounded tokens should contribute less to the state update.}
FGIP performs this state-conditioned token refinement, transforming the selected
sub-instruction into a visually grounded instruction-state update whose
internal emphasis adapts to the agent’s moment-to-moment perceptual
state.

\paragraph{Instruction Refiner}
\revisee{Let $S_{t-1}$ denote the instruction state from the previous navigation step, and let
$T_{k^\ast}$ denote the token indices corresponding to the sub-instruction selected by CGIP.}
FGIP first performs token-level visual grounding to determine which words in
$T_{k^\ast}$ are semantically active under the current observation $V_t$.
Multi-head cross-attention integrates linguistic and visual features:
\begin{equation}
\revisee{\tilde{T} = \mathrm{CrossAttn}(S_{t-1}[T_{k^\ast}],\, V_t),}
\end{equation}
producing visually modulated token embeddings.
\revisee{Here, $S_{t-1}[T_{k^\ast}] \in \mathbb{R}^{|T_{k^\ast}| \times d}$ denotes the token states
at the positions selected by CGIP and serves as the query, while the panoramic visual feature
$V_t$ provides the keys and values.}
This operation allows
tokens whose visual referents appear in the scene (e.g., ``sink'',
``hallway'', ``left'') to receive strong activation, while tokens lacking
visual grounding are naturally suppressed.

\revisee{Although cross-attention establishes token--region alignment, the internal
structure of the selected sub-instruction (e.g., modifier relations, phrase
boundaries, and compositional semantics) must also be preserved. We therefore
apply a lightweight Transformer encoder to the selected token states:
\begin{equation}
\hat{R}_t = \mathrm{Trans}(\tilde{T}),
\end{equation}
where $\hat{R}_t \in \mathbb{R}^{|T_{k^\ast}| \times d}$ denotes the refined
representation of the selected sub-instruction after local contextual modeling.
To integrate this refined segment back into the instruction state, we
scatter it to the original token positions:
\begin{equation}
R_t = \mathrm{Scatter}(\hat{R}_t, T_{k^\ast}),
\end{equation}
where $R_t \in \mathbb{R}^{L \times d}$ is the full-state representation whose
entries are updated only at the selected token positions and remain unchanged
elsewhere.
This refinement branch is optimized jointly with the downstream navigation objective, so that gradients from action prediction encourage the two-step update to produce a representation that jointly captures fine-grained perceptual grounding and the internal linguistic coherence of the selected sub-instruction.}

\paragraph{State Update via Gated Fusion}
FGIP incorporates the refined sub-instruction representation into the
instruction state using a gated residual update:
\begin{equation}
\revisee{S_t = S_{t-1} + g \odot (R_t - S_{t-1}),}
\end{equation}
\revisee{where $g \in (0,1)^{L \times d}$ is a learnable element-wise gate that adaptively controls how much the refined representation updates each token state. In implementation, we compute $g$ token-wise from the previous instruction state and the refined representation:
\begin{equation}
\label{eq:gate}
g=\sigma\!\left(\mathrm{MLP}_g\!\left([\mathrm{LN}(S_{t-1});\,\mathrm{LN}(R_t)]\right)\right),
\end{equation}
where $[\cdot;\cdot]$ denotes concatenation along the feature dimension, $\mathrm{LN}(\cdot)$ is LayerNorm, and $\mathrm{MLP}_g$ is a two-layer MLP.}
\revisee{Intuitively, tokens with strong perceptual support receive larger updates, while tokens with weak or ambiguous evidence remain close to their previous values.}

This gated fusion mechanism prevents semantic drift, maintains stability
across navigation steps, and ensures that the instruction state evolves only
when justified by the current observation. \revisee{Through this refinement-and-update pipeline, FGIP produces a viewpoint-aligned, state-conditioned instruction representation for downstream navigation decisions.}

\subsection{Training Objective}

S-EGIU is trained end-to-end using a hybrid imitation learning (IL) and
reinforcement learning (RL) objective. IL provides short-horizon
supervision that stabilizes the instruction--state update process, while RL
optimizes long-horizon navigation performance through interaction with the
environment.

\paragraph{Imitation Learning}
IL supervises decision making using teacher-forced cross-entropy.
\revisee{Under the straight-through clause routing described in CGIP, gradients from teacher-forced action prediction propagate through both FGIP and the clause-relevance distribution in CGIP, thereby providing end-to-end supervision for coarse sub-instruction activation and fine-grained token grounding.}
\revisee{This encourages the coarse-to-fine instruction state to remain consistent with the oracle trajectory and the agent's perceptual state.}

\paragraph{Reinforcement Learning}
RL is implemented using an Advantage Actor--Critic (A2C) objective, which
encourages policies that yield high navigation rewards. \revisee{The instruction state produced by CGIP and FGIP serves as part of the policy input, allowing RL to further optimize how instruction semantics evolve under long-horizon interaction beyond teacher-forced demonstrations.}

\paragraph{Overall Loss}
The model is optimized with a weighted combination of the IL and RL terms:
\begin{equation}
\mathcal{L} = \mathcal{L}^{RL} + \lambda\, \mathcal{L}^{IL},
\end{equation}
where $\lambda$ controls the strength of the supervised signal.
This hybrid training objective ensures both stability during early learning
and strong performance in long-horizon, dynamically grounded navigation.

\section{Experiments}
\revisee{We evaluate S-EGIU on three widely used VLN benchmarks: R2R, SOON, and REVERIE. Together, they cover short-horizon navigation, long compositional instructions, and object-centric grounding.}  \revisee{These datasets provide complementary challenges for assessing state-conditioned instruction modeling under diverse navigation settings, including evolving semantic relevance, viewpoint dependence, and fine-grained visual grounding.}
\revisee{All experiments follow standard VLN evaluation protocols and compare against representative and competitive prior VLN methods.}
\revisee{For several mechanism families not originally reported under our REVERIE protocol, we additionally provide controlled DUET-based plug-in comparisons to contextualize representative mechanism families under the same backbone.}

\subsection{Datasets}

R2R~\cite{anderson2018vision} contains short, concise navigation instructions paired with relatively simple trajectories.  
It primarily evaluates fine-grained grounding ability and generalization to unseen environments, \revisee{providing a clean setting to assess instruction grounding and generalization under short-horizon semantic transitions.}

SOON~\cite{zhu2021soon} features long-horizon navigation trajectories with multi-step, compositional instructions.  
Semantic relevance shifts significantly as the agent progresses, \revisee{making SOON a challenging benchmark for evaluating instruction modeling under long-horizon perceptual and semantic changes.}

REVERIE~\cite{qi2020reverie} augments navigation with object-centric grounding: the agent must localize target objects specified in the instruction after reaching the goal region.  
This requires precise token-level grounding and perceptual disambiguation, \revisee{making it a strong testbed for fine-grained instruction grounding and perceptual disambiguation.}

\subsection{Evaluation Metrics}

We adopt standard VLN evaluation metrics to assess both navigation success and path efficiency.  
Success Rate (SR) measures whether the agent stops within the target region, while Oracle Success Rate (OSR) reports whether the agent reaches the success radius at any point during the trajectory.  
Trajectory Length (TL) evaluates the path length, and 
Success weighted by Path Length (SPL) combines success with path optimality.

\revisee{For SOON, we additionally report Relative Goal SPL (RGSPL), which reflects goal proximity while penalizing inefficient trajectories.}
\revisee{Together, these metrics provide a comprehensive evaluation of navigation success, path efficiency, and progress toward the goal.}

\subsection{Implementation Details}

\revisee{We adopt DUET~\cite{chen2022think} as the base model because its modular and reproducible architecture facilitates controlled evaluation of the proposed state-conditioned instruction updates. Compared with larger foundation-model-based systems, DUET provides a cleaner setting for isolating mechanism-level contributions in ablations and comparative studies.}

All experiments are implemented in PyTorch and trained on NVIDIA RTX 3090 GPUs. We optimize S-EGIU using Adam with a learning rate of $1\times10^{-5}$ for 20k steps.  
Early stopping is applied based on validation SPL, and the final model is chosen according to the best unseen-validation performance. \revisee{Unless otherwise stated, we follow DUET's original dataset-specific training recipe and keep the optimization setting aligned with the corresponding DUET baseline for fair comparison.}
\revisee{Following the standard DUET setting, we use pre-extracted frozen ViT-B/16 panoramic visual tokens as the perceptual input $V_t$ for the coarse-to-fine instruction-state updates.}
All experiments follow standard VLN evaluation protocols to ensure fair and consistent comparison with previous methods.

\revisee{\textbf{Controlled plug-in baselines.}
To contextualize the proposed state-conditioned instruction updates against representative mechanism families in recent VLN research, we implement controlled plug-ins within the same DUET backbone~\cite{chen2022think}, inspired by compact memory, exploration/backtracking, retrieval memory, planning priors, and coarse-to-fine grounding, following MapNav, E2BA, VLMaps, iPPD, and C2F-VG, respectively~\cite{zhang2025mapnav,shi2025e2ba,huang2023vlmaps,wang2025ippd,mi2024c2f}. Since these methods are not all originally reported under the exact REVERIE protocol used in this paper, and their full systems may introduce additional modules or assumptions that are not directly comparable, we distill only the core mechanism of each family into the same DUET framework for a controlled comparison. All plug-ins therefore share the same perceptual input (frozen ViT-B/16 panoramic tokens), training schedule, and evaluation scripts as DUET, while excluding additional system assumptions such as external mapping, external planners, or extra supervision. The full implementation scope and default settings are provided in the supplementary material.}

\subsection{Main Results}
\revisee{We first report the main benchmark results on REVERIE, R2R, and SOON in Tables~\ref{tab:reverie_results}, \ref{tab:r2r_results}, and \ref{tab:soon_results}, respectively.}
\revisee{We then provide two additional analyses to further contextualize the proposed mechanism: a controlled DUET plug-in comparison on REVERIE and a backbone transfer study on R2R.}
\revisee{Definitions of the controlled plug-ins, their default settings, the full REVERIE plug-in results, and the extended backbone transfer results are provided in the supplementary material.}

\textbf{Results on REVERIE.}
\revisee{Table~\ref{tab:reverie_results} shows that S-EGIU consistently improves trajectory efficiency over strong baselines on the challenging object-centric REVERIE benchmark.
On Val Unseen, S-EGIU improves SR (48.37\%) and SPL (35.70\%) over DUET (46.98\% SR, 33.73\% SPL), indicating more accurate instruction-conditioned decisions under unseen environments.
On Test Unseen, S-EGIU achieves the best SPL (38.74\%), demonstrating more efficient navigation while maintaining competitive OSR/SR compared with DUET.}

\revisee{\textbf{Controlled plug-in comparison on REVERIE.}}
\revisee{Table~\ref{tab:reverie_results} shows that, under the same DUET backbone and matched training/evaluation protocol, different mechanism families yield clearly different gains. In particular, planning-prior and coarse-to-fine grounding plug-ins are more effective than compact-memory or retrieval-style variants on unseen splits. Compared with these controlled plug-ins, our method achieves the best SPL on both Val Unseen and Test Unseen (35.70 and 38.74), while also reducing trajectory length from 22.11 to 21.06 and from 21.59 to 17.53, respectively. These results suggest that, under a fixed backbone, the main benefit of our method lies not in adding an auxiliary mechanism alone, but in improving decision efficiency through state-conditioned instruction updates.}

\begin{table*}[t]
	\centering
\caption{\revisee{REVERIE results on Val Seen, Val Unseen, and Test Unseen. Beyond standard baselines, we additionally report five {controlled plug-in} variants instantiated in the same DUET backbone~\cite{chen2022think}, corresponding to representative mechanism families inspired by MapNav~\cite{zhang2025mapnav}, E2BA~\cite{shi2025e2ba}, VLMaps~\cite{huang2023vlmaps}, iPPD~\cite{wang2025ippd}, and C2F-VG~\cite{mi2024c2f}. Methods marked with $^{\dagger}$ are our distilled mechanism-level plug-ins evaluated under the REVERIE protocol, rather than the original full systems reported in their respective papers; detailed implementation scope and default settings are provided in the supplementary material.}}
\label{tab:reverie_results}
	\resizebox{\textwidth}{!}{
		\begin{tabular}{lcccc|cccc|cccc}
			\toprule
			& \multicolumn{4}{c}{\textbf{Val Seen}} & \multicolumn{4}{c}{\textbf{Val Unseen}} & \multicolumn{4}{c}{\textbf{Test Unseen}} \\
			\cmidrule(lr){2-5} \cmidrule(lr){6-9} \cmidrule(lr){10-13}
			\textbf{Model} & \textbf{TL} (↓) & \textbf{OSR} (↑) & \textbf{SR} (↑) & \textbf{SPL} (↑) & \textbf{TL} (↓) & \textbf{OSR} (↑) & \textbf{SR} (↑) & \textbf{SPL} (↑) & \textbf{TL} (↓) & \textbf{OSR} (↑) & \textbf{SR} (↑) & \textbf{SPL} (↑) \\
			\midrule
			Seq2Seq \cite{anderson2018vision} & 12.88 & 35.70 & 29.59 & 24.01 & 11.07 & 8.07 & 4.20 & 2.84 & 10.89 & 6.88 & 3.99 & 3.09 \\
			SMNA \cite{ma2019self} & \textbf{7.54} & 43.29 & 41.25 & 39.61 & \textbf{9.07} & 11.28 & 8.15 & 6.44 & \textbf{9.23} & 8.39 & 5.80 & 4.53 \\
			RecBERT \cite{hong2021vln} & 13.44 & 53.90 & 51.79 & 47.96 & 16.78 & 35.02 & 30.67 & 24.90 & 15.86 & 32.91 & 29.61 & 23.99 \\
			HAMT \cite{chen2021history} & 12.79 & 47.65 & 43.29 & 40.19 & 14.08 & 36.84 & 32.95 & 30.20 & 13.62 & 33.41 & 30.40 & 26.67 \\
			DUET \cite{chen2022think} & 13.86 & 73.86 & 71.75 & 63.94 & 22.11 & 51.07 & 46.98 & 33.73 & 21.59 & \textbf{59.61} & \textbf{52.51} & 36.06 \\
            \midrule
\revisee{MapNav$^{\dagger}$ \cite{zhang2025mapnav}} & \revisee{13.28} & \revisee{71.75} & \revisee{70.13} & \revisee{63.16} & \revisee{21.15} & \revisee{50.33} & \revisee{47.12} & \revisee{33.76} & \revisee{19.42} & \revisee{56.21} & \revisee{51.86} & \revisee{36.58} \\
\revisee{E2BA$^{\dagger}$ \cite{shi2025e2ba}} & \revisee{12.68} & \revisee{70.91} & \revisee{69.78} & \revisee{63.57} & \revisee{20.74} & \revisee{49.84} & \revisee{45.98} & \revisee{34.68} & \revisee{18.97} & \revisee{55.63} & \revisee{50.42} & \revisee{37.11} \\
\revisee{VLMaps$^{\dagger}$ \cite{huang2023vlmaps}} & \revisee{14.23} & \revisee{73.44} & \revisee{71.40} & \revisee{62.57} & \revisee{22.30} & \revisee{51.26} & \revisee{47.03} & \revisee{34.10} & \revisee{20.41} & \revisee{56.84} & \revisee{51.79} & \revisee{36.72} \\
\revisee{iPPD$^{\dagger}$ \cite{wang2025ippd}} & \revisee{13.91} & \revisee{74.70} & \revisee{72.17} & \revisee{63.76} & \revisee{22.37} & \revisee{53.22} & \revisee{48.99} & \revisee{34.81} & \revisee{20.26} & \revisee{58.11} & \revisee{53.36} & \revisee{37.48} \\
\revisee{C2F-VG$^{\dagger}$ \cite{mi2024c2f}} & \revisee{13.41} & \revisee{74.42} & \revisee{72.61} & \revisee{64.53} & \revisee{21.10} & \revisee{52.43} & \revisee{48.31} & \revisee{34.83} & \revisee{19.28} & \revisee{57.66} & \revisee{52.91} & \revisee{37.63} \\
            \midrule
            \textbf{Ours} & 11.93 & \textbf{76.32} & \textbf{74.91} & \textbf{71.05} & 21.06 & {52.15} & {48.37} & \textbf{35.70} & 17.53 & 54.63 & 51.92 & \textbf{38.74} \\
			\bottomrule
	\end{tabular}}
\end{table*}

\textbf{Results on R2R.}
Table~\ref{tab:r2r_results} presents results on R2R, which focuses on concise instruction-following.
On Val Unseen, S-EGIU improves SR (74\%) and SPL (61\%) over DUET (72\% SR, 60\% SPL), while reducing trajectory length (13.33 vs.\ 13.94).
Navigation error remains comparable (3.37\,m vs.\ 3.31\,m), suggesting that the main benefit on R2R lies in more efficient execution and a higher success rate rather than a uniformly lower endpoint error.
On Test Unseen, S-EGIU achieves the highest SPL (60\%) and competitive SR (68\%), indicating improved path efficiency with competitive generalization in unseen environments.

\begin{table*}[t]
	\centering
	\caption{Performance comparison on the R2R dataset for the Val Unseen and Test Unseen splits.}
	\label{tab:r2r_results}
	\begingroup
	\setlength{\tabcolsep}{10pt}
	\resizebox{\textwidth}{!}{
		\begin{tabular}{lcccc|cccc}
			\toprule
			& \multicolumn{4}{c}{\textbf{Val Unseen}} & \multicolumn{4}{c}{\textbf{Test Unseen}} \\
			\cmidrule(lr){2-5} \cmidrule(lr){6-9}
			\textbf{Model} & \textbf{TL} (↓) & \textbf{NE} (↓) & \textbf{SR} (↑) & \textbf{SPL} (↑) & \textbf{TL} (↓) & \textbf{NE} (↓) & \textbf{SR} (↑) & \textbf{SPL} (↑) \\
			\midrule
			Seq2Seq \cite{anderson2018vision} & \textbf{8.39} & 7.81 & 22 & - & \textbf{8.13} & 7.85 & 20 & 18 \\
			RecBERT \cite{hong2021vln} & 12.01 & 3.93 & 63 & 57 & 12.35 & 4.09 & 63 & 57 \\
			HAMT \cite{chen2021history} & 11.46 & 2.29 & 66 & 61 & 12.27 & 3.93 & 65 & 60 \\
			DUET \cite{chen2022think} & 13.94 & \textbf{3.31} & 72 & 60 & 14.73 & 3.65 & \textbf{69} & 59 \\
			\textbf{Ours} & 13.33 & 3.37 & \textbf{74} & \textbf{61} & 13.96 & \textbf{3.43} & 68 & \textbf{60} \\
			\bottomrule
	\end{tabular}}
	\endgroup
\end{table*}

\textbf{Results on SOON.}
Table~\ref{tab:soon_results} reports results on SOON, a long-horizon benchmark characterized by multiple sub-instructions and frequent context transitions. On the Val Unseen split, S-EGIU attains comparable SR (35.69\%) to DUET while
achieving higher SPL (24.95\% vs.\ 22.58\%) and a shorter TL (33.10 vs.\ 36.20), indicating more efficient path execution.  
A similar pattern holds on Test Unseen, where S-EGIU improves SPL (23.61\%) and substantially reduces TL (37.37 vs.\ 41.83). \revisee{These results suggest that the proposed state-conditioned instruction updates are particularly beneficial for path efficiency under SOON's long, semantically layered instructions.}

\begin{table*}[t]
	\centering
	\caption{Performance comparison on the SOON dataset for Val Unseen and Test Unseen splits.}
	\label{tab:soon_results}
	\resizebox{\textwidth}{!}{
		\begin{tabular}{lccccc|ccccc}
			\toprule
			& \multicolumn{5}{c}{\textbf{Val Unseen}} & \multicolumn{5}{c}{\textbf{Test Unseen}} \\
			\cmidrule(lr){2-6} \cmidrule(lr){7-11}
			\textbf{Model} & \textbf{TL} (↓) & \textbf{OSR} (↑) & \textbf{SR} (↑) & \textbf{SPL} (↑) & \textbf{RGSPL} (↑) & \textbf{TL} (↓) & \textbf{OSR} (↑) & \textbf{SR} (↑) & \textbf{SPL} (↑) & \textbf{RGSPL} (↑) \\
			\midrule
			GBE \cite{zhu2021soon} & \textbf{28.96} & 28.54 & 19.52 & 13.34 & 1.16 & \textbf{27.88} & 21.45 & 12.90 & 9.23 & 0.45 \\
			DUET \cite{chen2022think} & 36.20 & \textbf{50.91} & \textbf{36.28} & 22.58 & \textbf{3.75} & 41.83 & \textbf{43.00} & \textbf{33.44} & 21.42 & \textbf{4.17} \\
			\textbf{Ours} & 33.10 & 47.17 & 35.69 & \textbf{24.95} & 3.47 & 37.37 & 40.82 & 31.76 & \textbf{23.61} & 3.95 \\
			\bottomrule
	\end{tabular}}
\end{table*}

\begin{table}[t]
\centering
\caption{\revisee{Generalization of S-EGIU across different backbones on R2R.
We report Val Seen and Val Unseen results under matched within-backbone protocols.
$\Delta$SR and $\Delta$SPL denote absolute gains over the corresponding baseline.}}
\label{tab:backbone_transfer_r2r}

\revisee{
\setlength{\tabcolsep}{3.8pt}
\resizebox{0.99\linewidth}{!}{
\begin{tabular}{lcccccc@{\hspace{8pt}}cccccc}
\toprule
& \multicolumn{6}{c}{\textbf{Val Seen}} & \multicolumn{6}{c}{\textbf{Val Unseen}} \\
\cmidrule(lr){2-7} \cmidrule(lr){8-13}
\textbf{Model}
& \textbf{TL} ($\downarrow$) & \textbf{NE} ($\downarrow$) & \textbf{SR} ($\uparrow$) & \textbf{$\Delta$SR} & \textbf{SPL} ($\uparrow$) & \textbf{$\Delta$SPL}
& \textbf{TL} ($\downarrow$) & \textbf{NE} ($\downarrow$) & \textbf{SR} ($\uparrow$) & \textbf{$\Delta$SR} & \textbf{SPL} ($\uparrow$) & \textbf{$\Delta$SPL} \\
\midrule
DUET\cite{chen2022think}
& 12.32 & 2.28 & 79 & -- & 73 & --
& 13.94 & 3.31 & 72 & -- & 60 & -- \\
DUET + S-EGIU
& 12.00 & 2.24 & 81 & +2 & 75 & +2
& 13.33 & 3.37 & 74 & +2 & 61 & +1 \\
\midrule
NavGPT2~\cite{zhou2024navgpt2}
& 12.44 & 2.97 & 73 & -- & 65 & --
& 12.81 & 3.33 & 70 & -- & 59 & -- \\
NavGPT2 + S-EGIU
& 12.05 & 2.88 & 75 & +2 & 67 & +2
& 12.30 & 3.22 & 72 & +2 & 60 & +1 \\
\midrule
HAMT\cite{chen2021history}
& 11.15 & 2.51 & 76 & -- & 72 & --
& 11.46 & 2.29 & 66 & -- & 61 & -- \\
HAMT + S-EGIU
& 10.95 & 2.42 & 77 & +1 & 73 & +1
& 11.20 & 2.24 & 67 & +1 & 62 & +1 \\
\midrule
VLN-GOAT\cite{wang2024causal}
& 11.90 & 1.79 & 84 & -- & 79 & --
& 12.40 & 2.40 & 78 & -- & 68 & -- \\
VLN-GOAT + S-EGIU
& 11.74 & 1.74 & 85 & +1 & 80 & +1
& 12.20 & 2.34 & 79 & +1 & 69 & +1 \\
\bottomrule
\end{tabular}}
}
\end{table}
\revisee{\textbf{Backbone transfer on R2R.}}
\revisee{Unlike Table~\ref{tab:r2r_results}, which compares against published methods on the standard benchmark splits, Table~\ref{tab:backbone_transfer_r2r} examines whether the proposed mechanism transfers consistently across heterogeneous backbones under matched within-backbone protocols.}
\revisee{Specifically, beyond DUET, we integrate S-EGIU into three additional representative backbones, namely NavGPT2, HAMT, and VLN-GOAT, and report results on R2R because its training and evaluation pipeline is comparatively standardized across codebases, enabling cleaner protocol matching for cross-backbone integration.}
\revisee{REVERIE is instead used in our main study to stress-test object-centric grounding, whereas its dataset-specific grounding heads and evaluation scripts could otherwise introduce additional confounds in cross-backbone comparisons.}

\revisee{Table~\ref{tab:backbone_transfer_r2r} shows that S-EGIU consistently improves SR and SPL across all tested backbones on both Val Seen and Val Unseen.}
\revisee{Averaged over the four backbones, the gains are +1.5 SR / +1.5 SPL on Val Seen and +1.5 SR / +1.0 SPL on Val Unseen.}
\revisee{Trajectory length is reduced in all eight comparisons, and NE is improved in seven of them, with only a minor degradation on DUET Val Unseen (3.31 to 3.37).}
\revisee{Although the absolute gains are moderate, their consistency across substantially different backbones provides direct evidence that the proposed state-conditioned instruction updates are transferable rather than DUET-specific.}

\revisee{In summary, S-EGIU consistently improves trajectory efficiency while maintaining competitive navigation success across diverse VLN benchmarks and heterogeneous backbones, supporting the effectiveness of state-conditioned instruction updates.}

\subsection{Ablation Study and Visualization}  

\textbf{Ablation Study.}
To assess the contribution of each component in S-EGIU, we conduct ablation experiments on REVERIE, evaluating three configurations: (1) CGIP-only, (2) FGIP-only, and (3) the full model with both CGIP and FGIP. Results on the Val Seen and Unseen splits are reported in Table~\ref{tab:ablation}.

\begin{table*}[t]
\centering
\caption{Ablation study on the REVERIE dataset, evaluating the impact of CGIP and FGIP.}
	\label{tab:ablation}
	\resizebox{\textwidth}{!}{
		\begin{tabular}{cc|cccc|cccc}
			\toprule
			& & \multicolumn{4}{c}{\textbf{Val Seen}} & \multicolumn{4}{c}{\textbf{Val Unseen}} \\
			\cmidrule(lr){3-6} \cmidrule(lr){7-10}
			\textbf{CGIP} & \textbf{FGIP} & \textbf{TL} (↓) & \textbf{OSR} (↑) & \textbf{SR} (↑) & \textbf{SPL} (↑) & \textbf{TL} (↓) & \textbf{OSR} (↑) & \textbf{SR} (↑) & \textbf{SPL} (↑) \\
			\midrule
			\ding{51} & \ding{55} & 10.83 & 74.28 & 72.96 & 70.78 & 22.07 & 51.71 & 47.47 & 34.11 \\
			\ding{55} & \ding{51} & 14.40 & 76.97 & 73.01 & 68.56 & 22.90 & 51.25 & 47.92 & 33.95 \\
			\ding{51} & \ding{51} & \textbf{11.93} & \textbf{76.32} & \textbf{74.91} & \textbf{71.05} & \textbf{21.06} & \textbf{52.15} & \textbf{48.37} & \textbf{35.70} \\
			\bottomrule
	\end{tabular}}
\end{table*}

\begin{figure*}[!t]
    \centering
    \includegraphics[width=\linewidth]{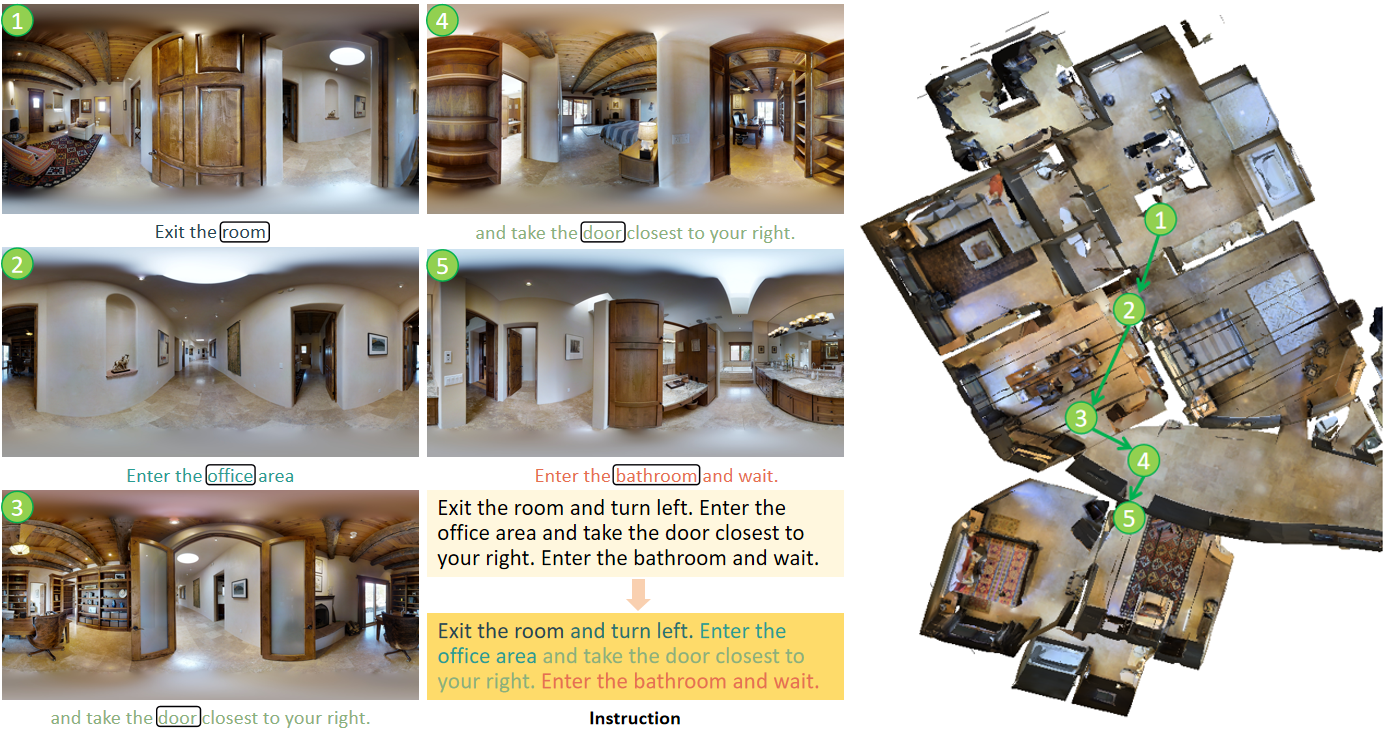}%
   
    \caption{Visualization of S-EGIU’s state-conditioned instruction understanding. Panoramic observations (left) reflect the agent’s evolving perceptual state, while the top-down map (right) shows the executed trajectory. The instruction is decomposed into color-coded sub-instructions, with perception-relevant tokens highlighted. This visualization illustrates how S-EGIU selects the perceptually valid sub-instruction and refines its token semantics to maintain alignment with the dynamic visual context.}
     \label{fig:vis}
\end{figure*}

\revisee{CGIP-only mainly improves trajectory efficiency, indicating that selecting the perceptually relevant sub-instruction helps reduce semantic ambiguity and supports more stable navigation decisions.}

\revisee{FGIP-only mainly improves success-oriented metrics among the single-component variants, suggesting that token-level refinement particularly benefits local grounding quality at each step.}

\revisee{Combining CGIP and FGIP yields the best overall performance, confirming that the two modules are complementary: CGIP provides a structured sub-instruction prior, while FGIP performs fine-grained semantic refinement within the active segment.}

\revisee{We focus this ablation on the two core modules while keeping all other settings unchanged. More fine-grained design variations, such as alternative gating or temporal weighting schemes, remain interesting directions for future study.}

\textbf{Visualization.}
\revisee{Fig.~\ref{fig:vis} qualitatively illustrates how S-EGIU updates instruction understanding in response to changing perceptual input.}
\revisee{The panoramic observations reflect the evolving perceptual state, the color-coded instruction segments indicate which sub-instruction is currently activated, and the highlighted tokens show how FGIP refines the active segment according to the current environment.}
\revisee{Together with the top-down trajectory, the figure demonstrates how state-conditioned instruction updates translate into more coherent and efficient navigation behavior.}

\revisee{Overall, the visualization illustrates how S-EGIU adaptively activates and refines instruction content in response to changing perceptual state.}

\begin{figure*}[t]
    \centering
    \includegraphics[width=\linewidth]{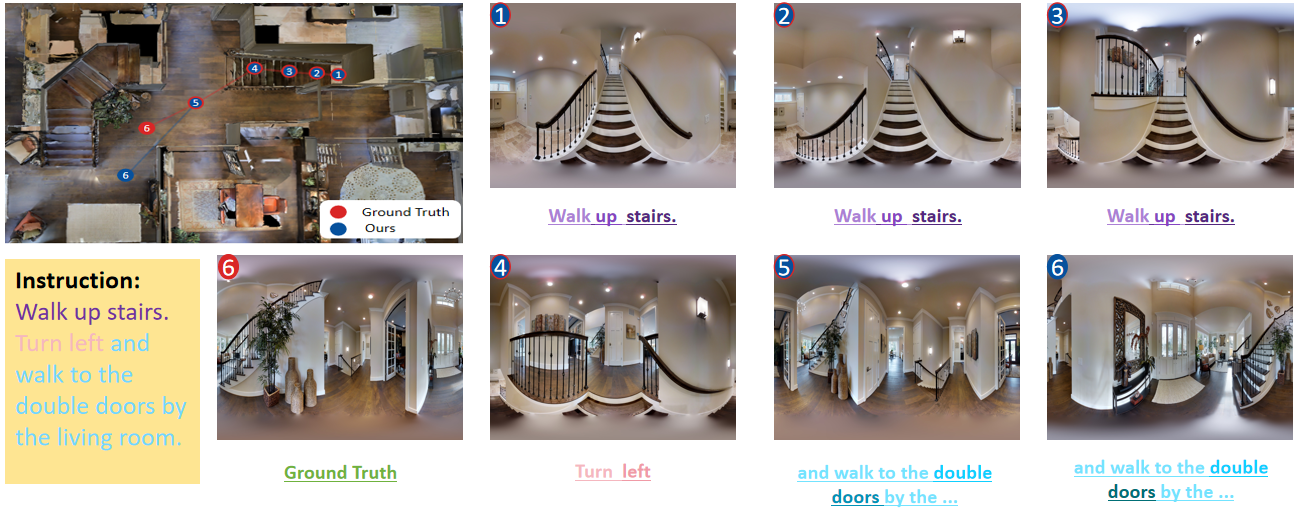}
    \caption{
    A representative failure case on the R2R dataset. 
    The agent (blue trajectory) follows the instruction correctly until the 
    final step, where it misidentifies the correct doorway 
    (ground-truth shown in red). 
    Colored and intensity-varied sub-instructions reflect the model’s 
    evolving internal instruction state and token importance.
    }
    \label{fig:failure_case}
\end{figure*}

\subsection{Error Analysis in Unseen Environments}

\revisee{Although S-EGIU achieves strong overall performance, its gains are generally smaller on unseen environments than on seen splits. Qualitative inspection suggests two plausible contributors to this gap, and the representative failure case below illustrates how such mismatches arise in practice.}

\revisee{First, unseen scenes introduce novel object layouts, visual appearances, and room topologies, which can disrupt the clause-selection stage in CGIP. When expected landmarks or spatial cues do not appear in the anticipated form or location, the model may activate a visually mismatched sub-instruction. Since subsequent refinement is conditioned on the selected clause, such early-stage selection errors can propagate to later updates.}

\revisee{Second, FGIP relies on panoramic visual cues to refine the selected sub-instruction at the token level. Under substantial cross-environment shifts in geometry, appearance, or lighting, the alignment between projected visual features and language tokens may become less reliable, leading to suboptimal refinement and less effective navigation decisions.}

\revisee{These observations suggest that improving robustness to cross-environment perceptual mismatch remains important for unseen generalization. Future extensions may incorporate domain-invariant priors or uncertainty-aware weighting to stabilize clause activation and refinement under visual distribution shift.}

\subsection{Failure Case Analysis}

Fig.~\ref{fig:failure_case} shows a representative failure case from R2R. 
The agent correctly follows the instruction for most of the trajectory but 
stops at an incorrect doorway in the final step.

To visualize the model’s internal reasoning, we color-code 
sub-instruction segments according to the active instruction state, with 
darker text indicating higher token importance. 
This provides an intuitive view of how S-EGIU transitions between segments 
and how token-level focus evolves with perception.
In this example, the model activates the appropriate sub-instruction but 
misinterprets the final visual cue, causing a misaligned state update and 
an incorrect stopping decision. 
\revisee{Such cases illustrate how small grounding errors near the goal can still affect the state-conditioned update process.}

\textbf{Recurring failure modes.}
\revisee{Qualitative inspection on R2R, SOON, and REVERIE reveals three common failure modes.}

\revisee{\textit{Ambiguous landmark discrimination.} In environments with repeated structures (e.g., similar doors or hallways), the model may activate the correct sub-instruction but still ground it to the wrong referent because the relevant visual cues are insufficiently distinctive.}

\revisee{\textit{Premature or delayed sub-instruction transitions.} When transitional cues are weak, partial, or occluded, CGIP may switch to the next sub-instruction too early or too late, leading to temporary misalignment between the instruction state and the current perceptual state.}

\revisee{\textit{Long-horizon drift in iterative updates.} Over extended trajectories, small grounding errors can accumulate through repeated FGIP updates, gradually shifting the instruction state away from the intended semantic focus, especially in multi-room environments.}

\revisee{These patterns suggest that stabilizing state-conditioned instruction updates under perceptual uncertainty remains a key challenge, and point to future directions such as uncertainty-aware grounding, transition regularization, and temporal smoothing.}

\section{Conclusion}

In this paper, we introduced S-EGIU, a framework that models instruction understanding as an evolving state entangled with the agent’s perceptual trajectory. By integrating coarse-grained sub-instruction selection with fine-grained, perception-guided token refinement, S-EGIU enables dynamic instruction–perception entanglement and prevents semantic drift throughout navigation.
Extensive experiments demonstrate that S-EGIU yields consistent efficiency gains across R2R, SOON, and REVERIE while maintaining competitive navigation success, including a notable +2.68\% SPL gain on REVERIE Test Unseen. These results highlight the importance of state-conditioned instruction semantics for robust embodied navigation.
\revisee{Limitations include the reliance on the quality of sub-instruction boundary refinement and the use of panoramic visual tokens as perceptual state, which may introduce sensitivity under severe viewpoint ambiguity or domain shift.}
Future work may explore richer temporal dynamics for instruction-state evolution and incorporate uncertainty-aware grounding to further strengthen robustness under ambiguous or visually complex conditions, enabling more adaptive and reliable navigation in real-world environments.

% \section{Example Appendix Section}


%%       \cite{<label>} ==> [1]

%%
% Example citation, See \cite{lamport94}.




% =========================
% Supplementary Materials (Final submission style: no \emph{}, no bold text)
% =========================
% =========================
% Supplementary Materials (Final submission version, abstract & paper-style)
% =========================
\newpage
\begingroup
% \color{red}
% \arrayrulecolor{red}

\section*{Supplementary Materials}
\label{sec:supp}

\setcounter{table}{0}
\renewcommand{\thetable}{S\arabic{table}}
\setcounter{figure}{0}
\renewcommand{\thefigure}{S\arabic{figure}}

\subsection*{S1. Purpose and Scope}
This supplementary material provides additional details for the two auxiliary analyses summarized in the main paper: the controlled plug-in comparison on REVERIE and the backbone transfer study on R2R~\cite{li2025regnav}.
Its purpose is to make the comparison protocol, implementation scope, default settings, and extended results transparent.

Specifically, this supplement serves three roles.
First, it formalizes the controlled plug-in protocol used to contextualize our method against representative mechanism families under a fixed DUET backbone.
Second, it clarifies the implementation scope and default settings of the five controlled DUET variants used in the REVERIE comparison.
Third, it reports extended result tables and reproducibility notes omitted from the main paper due to space limitations.

Unless otherwise stated, all controlled variants share the same perceptual input, training schedule, and evaluation scripts as the original DUET backbone~\cite{wu2024cr,fu2019recognition,fu2024understanding}.
No additional sensors, external mapping modules, external planning systems, or extra supervision signals are introduced.
Accordingly, this supplement should be read as supporting evidence for mechanism-level comparison under a fixed backbone and transferability across backbones~\cite{li2024camera}.

\subsection*{S2. Controlled Plug-in Protocol}
\label{sec:supp_rationale}
The main paper studies {state-conditioned instruction updates}, where the instruction is treated as a state that is dynamically updated through perception via CGIP and FGIP~\cite{wu2025event,liu2024semantic,liu2025mind}.
To contextualize this mechanism, we compare against five representative improvement families frequently associated with recent VLN gains, including compact memory, exploration/backtracking, retrieval memory, planning priors, and coarse-to-fine grounding, following MapNav, E2BA, VLMaps, iPPD, and C2F-VG, respectively~\cite{zhang2025mapnav,shi2025e2ba,huang2023vlmaps,wang2025ippd,mi2024c2f}.

A direct full-system comparison is often difficult to interpret because many prior methods differ not only in their core mechanism, but also in backbone architecture, auxiliary modules, sensor assumptions, training pipelines, or external map/planner dependencies.
Moreover, several methods are not originally reported under the exact REVERIE protocol used in this paper, and their full systems may include additional components that are not directly comparable under a controlled setting.
To reduce these confounds, we adopt a controlled plug-in protocol: each mechanism family is distilled into a lightweight DUET-based variant and trained/evaluated under the same input setting, training schedule, and evaluation scripts as DUET.

For naming consistency with the main paper, we refer to each controlled variant using the corresponding representative method family, namely MapNav, E2BA, VLMaps, iPPD, and C2F-VG.
These names should be interpreted at the level of {mechanism families} rather than as claims of one-to-one functional equivalence with the original full systems.
Accordingly, the variants reported here are DUET-based controlled distillations for mechanism-level comparison rather than full re-implementations of the published methods.

\subsection*{S3. Controlled Plug-in Instantiations}
\label{sec:supp_plugins}
All controlled DUET variants preserve DUET's original input space, action space, and training objectives.
They are implemented as lightweight insertions at one of three locations:
(i) representation conditioning before action scoring,
(ii) action-score biasing before score fusion, or
(iii) object-score biasing for object-centric grounding on REVERIE~\cite{qi2026patchcue}.
Unless otherwise noted, all variants are trained end-to-end together with the original DUET losses and do not rely on extra supervision.

\noindent\textbf{Controlled comparison rule.}
All plug-ins are implemented within DUET and share identical inputs (frozen ViT-B/16 panoramic tokens), training schedule, and evaluation scripts.
We exclude additional system assumptions such as extra sensors, external SLAM/3D mapping, external planners, or extra training data, so that the comparison remains controlled at the mechanism level.

\begin{table*}[t]
\centering
\caption{Controlled plug-in baselines: implementation scope summary. Each variant is a DUET-integrated controlled distillation of a representative mechanism family; details are given in Sec.~S3.1--S3.6.}
\label{tab:plugin_fidelity}
\resizebox{\textwidth}{!}{
\begin{tabular}{l p{6.3cm} p{5.4cm} l}
\toprule
\textbf{Family} & \textbf{DUET plug-in (what is added)} & \textbf{Excluded for control} & \textbf{Pointer} \\
\midrule
MapNav$^{\dagger}$~\cite{zhang2025mapnav} & \textbf{MemTok:} fixed-length memory tokens updated each step and injected before action scoring & no extra sensors; no external SLAM/3D mapping; no extra pretraining data & Sec.~S3.1 \\
E2BA$^{\dagger}$~\cite{shi2025e2ba} & \textbf{Backtrack:} exploration/return bias added to action scores using an uncertainty proxy & no external exploration engine/planner; no particle filter; no external scoring modules & Sec.~S3.2 \\
VLMaps$^{\dagger}$~\cite{huang2023vlmaps} & \textbf{KV-Cache:} bounded language-queryable retrieval cache injected as context & no explicit 3D reconstruction/SLAM; no depth/odometry; no metric-grid fusion & Sec.~S3.3 \\
iPPD$^{\dagger}$~\cite{wang2025ippd} & \textbf{PathPrior:} instruction-conditioned prior added to global node/action scores & no explicit map construction; no external planner; no extra supervision beyond DUET & Sec.~S3.4 \\
C2F-VG$^{\dagger}$~\cite{mi2024c2f} & \textbf{C2F-Ground:} coarse-to-fine grounding prior injected into object scores (REVERIE) & no extra object annotations; no additional grounding datasets; no extra detector training & Sec.~S3.5 \\
\bottomrule
\end{tabular}}
\vspace{0.2em}

\footnotesize{$^{\dagger}$ denotes DUET-integrated controlled distillations for controlled mechanism-level comparison, not full-system re-implementations.}
\end{table*}

\subsubsection*{S3.1 MapNav-style Control (MemTok): Compact Memory}
This control tests whether bounded history summarization alone can explain gains without explicitly updating instruction semantics.
We maintain a fixed-length set of memory tokens updated per step from DUET internal summaries and inject them as additional context before action scoring.

\subsubsection*{S3.2 E2BA-style Control (Backtrack): Exploration and Return Bias}
This control isolates gains from action-selection bias (exploration and return behaviors) under a fixed backbone.
We add a lightweight bias to action scores that strengthens return preference when the policy becomes uncertain, without changing DUET's action space or introducing an external planner.

\subsubsection*{S3.3 VLMaps-style Control (KV-Cache): Retrieval Memory}
This control isolates retrieval-based contextualization from explicit instruction-state updates.
We maintain a bounded language-queryable key--value cache over steps; the instruction summary retrieves a context vector that is injected before action scoring.

\subsubsection*{S3.4 iPPD-style Control (PathPrior): Global Planning Prior}
This control tests whether an instruction-conditioned global inclination over candidate nodes can account for gains.
We compute a similarity-based prior between the instruction summary and candidate node representations and add it to DUET global action scores (masked on invalid actions), without introducing external map construction or planners.

\subsubsection*{S3.5 C2F-VG-style Control (C2F-Ground): Coarse-to-fine Grounding}
This control isolates object grounding improvements on REVERIE from instruction-state modeling~\cite{qi2026patchcue}.
We select top-$K$ objects using a coarse relevance score and refine only this subset via lightweight cross-modal matching, injecting the resulting grounding prior into DUET object scores.

\subsubsection*{S3.6 Minimal Implementation Notes (shared across plug-ins)}
\label{sec:supp_plugin_notes}
We use bounded modules to keep the comparison controlled: MemTok uses $K{=}16$ memory tokens with EMA smoothing $\alpha{=}0.9$; KV-Cache has capacity 64 with oldest-entry eviction; C2F-Ground uses top-$K{=}5$ for coarse filtering.
Backtrack uses action-distribution uncertainty (normalized entropy) as a confidence proxy to modulate return bias, and PathPrior uses similarity-based node priors with weight 0.3 and temperature $T_p{=}1.0$; STOP is not biased by default.

\subsection*{S4. Default Settings and Tuning Policy}
\label{sec:supp_hparams}
Table~\ref{tab:supp_plugin_hparams} summarizes the default settings used in the controlled plug-in study.
We intentionally keep all controlled variants lightweight and restrict the hyperparameter search budget to the same validation-level tuning procedure, so that the comparison emphasizes mechanism differences rather than aggressive variant-specific optimization~\cite{wang2023closing,fu2023and,fu2023breaking}.

\begin{table}[t]
\centering
\caption{Default settings for the controlled DUET plug-in study.}
\label{tab:supp_plugin_hparams}
\resizebox{0.98\linewidth}{!}{
\begin{tabular}{lccc}
\toprule
Variant & Main setting & Default & Injection level \\
\midrule
MapNav-style (MemTok) & memory size and smoothing & 16 tokens, smoothing 0.9 & representation \\
E2BA-style (Backtrack) & exploration and return weights & 0.2 and 0.5 & action scores \\
VLMaps-style (KV-Cache) & cache capacity & 64 entries & representation \\
iPPD-style (PathPrior) & prior strength and temperature & 0.3 and 1.0 & global scores \\
C2F-VG-style (C2F-Ground) & coarse top-K and bias strength & 5 and 0.5 & object scores \\
\bottomrule
\end{tabular}}
\end{table}

All DUET-based controlled variants are selected using the same validation split and checkpoint-selection criterion as the base DUET model.
We apply only lightweight validation-level tuning for each variant, using the same search budget and the same model-selection rule across all plug-ins.
The goal of this procedure is not to maximize each variant independently, but to keep the comparison stable and comparable across mechanism families.

\subsection*{S5. Extended Results}
\label{sec:supp_results}

\subsubsection*{S5.1 Controlled DUET Variants on REVERIE}
\label{sec:supp_reverie_plugins}
Table~\ref{tab:supp_reverie_plugins} expands the controlled comparison summarized in the main paper by reporting the full REVERIE results for DUET, its controlled variants, and our full model under the same protocol.
These results provide a mechanism-level reference for how much improvement can be obtained by memory, exploration, retrieval, planning, or grounding enhancements alone under a fixed backbone.

\begin{table*}[t]
\centering
\caption{REVERIE results of DUET, controlled DUET variants, and our full model under the same protocol.
Variants marked with $^{\dagger}$ are DUET-integrated controlled distillations named after representative method families for consistency with the main paper, rather than full-system re-implementations.}
\label{tab:supp_reverie_plugins}
\resizebox{\textwidth}{!}{
\begin{tabular}{lcccc|cccc|cccc}
\toprule
& \multicolumn{4}{c}{Val Seen} & \multicolumn{4}{c}{Val Unseen} & \multicolumn{4}{c}{Test Unseen} \\
\cmidrule(lr){2-5} \cmidrule(lr){6-9} \cmidrule(lr){10-13}
Model & TL (↓) & OSR (↑) & SR (↑) & SPL (↑) &
TL (↓) & OSR (↑) & SR (↑) & SPL (↑) &
TL (↓) & OSR (↑) & SR (↑) & SPL (↑) \\
\midrule
DUET & 13.86 & 73.86 & 71.75 & 63.94 & 22.11 & 51.07 & 46.98 & 33.73 & 21.59 & 59.61 & 52.51 & 36.06 \\
MapNav$^{\dagger}$ & 13.28 & 71.75 & 70.13 & 63.16 & 21.15 & 50.33 & 47.12 & 33.76 & 19.42 & 56.21 & 51.86 & 36.58 \\
E2BA$^{\dagger}$ & 12.68 & 70.91 & 69.78 & 63.57 & 20.74 & 49.84 & 45.98 & 34.68 & 18.97 & 55.63 & 50.42 & 37.11 \\
VLMaps$^{\dagger}$ & 14.23 & 73.44 & 71.40 & 62.57 & 22.30 & 51.26 & 47.03 & 34.10 & 20.41 & 56.84 & 51.79 & 36.72 \\
iPPD$^{\dagger}$ & 13.91 & 74.70 & 72.17 & 63.76 & 22.37 & 53.22 & 48.99 & 34.81 & 20.26 & 58.11 & 53.36 & 37.48 \\
C2F-VG$^{\dagger}$ & 13.41 & 74.42 & 72.61 & 64.53 & 21.10 & 52.43 & 48.31 & 34.83 & 19.28 & 57.66 & 52.91 & 37.63 \\
Ours (S-EGIU) & 11.93 & 76.32 & 74.91 & 71.05 & 21.06 & 52.15 & 48.37 & 35.70 & 17.53 & 54.63 & 51.92 & 38.74 \\
\bottomrule
\end{tabular}}
\end{table*}

Table~\ref{tab:supp_reverie_plugins} shows that different controlled variants improve different aspects of DUET under the same backbone.
On Val Unseen, iPPD and C2F-VG improve SR from 46.98 to 48.99 and 48.31, respectively, and improve SPL from 33.73 to 34.81 and 34.83, indicating that planning-prior and coarse-to-fine grounding controls provide relatively stronger gains than compact-memory or retrieval-style variants under this fixed setting.
By contrast, MapNav and VLMaps mainly offer modest gains, while E2BA improves SPL but slightly reduces SR on Val Unseen.
Compared with all controlled plug-ins, S-EGIU yields the strongest efficiency-oriented gain, achieving the best SPL on both Val Unseen and Test Unseen (35.70 and 38.74) while reducing trajectory length from 22.11 to 21.06 and from 21.59 to 17.53, respectively.
Overall, these results suggest that the benefit of our method does not arise from merely adding a memory, retrieval, planning, or grounding bias in isolation, but from explicitly updating instruction semantics as a state~\cite{liu2024semantic,liu2025mind}.

\subsubsection*{S5.2 Backbone Transfer on R2R}
\label{sec:supp_r2r_transfer}
Table~\ref{tab:supp_backbone_transfer_r2r} expands the backbone generalization analysis summarized in the main paper.
Unlike the main R2R benchmark table, which compares against published methods on the standard splits, this study examines whether the proposed mechanism transfers consistently across heterogeneous backbones under matched within-backbone protocols.
Specifically, beyond DUET, we integrate S-EGIU into three additional representative backbones, namely NavGPT2, HAMT, and VLN-GOAT, and evaluate them on R2R.
We use R2R for this transfer study because its training and evaluation pipeline is comparatively standardized across codebases, enabling cleaner protocol matching for cross-backbone integration.
REVERIE is instead used in the main paper to stress-test object-centric grounding, whereas its dataset-specific grounding heads and evaluation scripts could otherwise introduce additional confounds in cross-backbone comparisons.

\begin{table*}[t]
\centering
\caption{Backbone transfer results on R2R under matched within-backbone protocols.
$\Delta$SR and $\Delta$SPL denote absolute gains over the corresponding baseline.}
\label{tab:supp_backbone_transfer_r2r}
\resizebox{\textwidth}{!}{
\begin{tabular}{lcccccc@{\hspace{8pt}}cccccc}
\toprule
& \multicolumn{6}{c}{\textbf{Val Seen}} & \multicolumn{6}{c}{\textbf{Val Unseen}} \\
\cmidrule(lr){2-7} \cmidrule(lr){8-13}
\textbf{Model}
& \textbf{TL} ($\downarrow$) & \textbf{NE} ($\downarrow$) & \textbf{SR} ($\uparrow$) & \textbf{$\Delta$SR} & \textbf{SPL} ($\uparrow$) & \textbf{$\Delta$SPL}
& \textbf{TL} ($\downarrow$) & \textbf{NE} ($\downarrow$) & \textbf{SR} ($\uparrow$) & \textbf{$\Delta$SR} & \textbf{SPL} ($\uparrow$) & \textbf{$\Delta$SPL} \\
\midrule
DUET
& 12.32 & 2.28 & 79 & -- & 73 & --
& 13.94 & 3.31 & 72 & -- & 60 & -- \\
DUET + S-EGIU
& 12.00 & 2.24 & 81 & +2 & 75 & +2
& 13.33 & 3.37 & 74 & +2 & 61 & +1 \\
\midrule
NavGPT2
& 12.44 & 2.97 & 73 & -- & 65 & --
& 12.81 & 3.33 & 70 & -- & 59 & -- \\
NavGPT2 + S-EGIU
& 12.05 & 2.88 & 75 & +2 & 67 & +2
& 12.30 & 3.22 & 72 & +2 & 60 & +1 \\
\midrule
HAMT
& 11.15 & 2.51 & 76 & -- & 72 & --
& 11.46 & 2.29 & 66 & -- & 61 & -- \\
HAMT + S-EGIU
& 10.95 & 2.42 & 77 & +1 & 73 & +1
& 11.20 & 2.24 & 67 & +1 & 62 & +1 \\
\midrule
VLN-GOAT
& 11.90 & 1.79 & 84 & -- & 79 & --
& 12.40 & 2.40 & 78 & -- & 68 & -- \\
VLN-GOAT + S-EGIU
& 11.74 & 1.74 & 85 & +1 & 80 & +1
& 12.20 & 2.34 & 79 & +1 & 69 & +1 \\
\bottomrule
\end{tabular}}
\end{table*}

Table~\ref{tab:supp_backbone_transfer_r2r} shows that S-EGIU consistently improves SR and SPL across all tested backbones on both Val Seen and Val Unseen.
Averaged over the four backbones, the gains are +1.5 SR / +1.5 SPL on Val Seen and +1.5 SR / +1.0 SPL on Val Unseen.
Trajectory length is reduced in all eight comparisons, and NE is improved in seven of them, with only a minor degradation on DUET Val Unseen (3.31 to 3.37).
Although the absolute gains are moderate, their consistency across substantially different backbones provides direct evidence that the proposed state-conditioned instruction updates are transferable rather than DUET-specific~\cite{li2024camera,wu2025event}.

\subsection*{S6. Reproducibility Notes}
\label{sec:supp_repro}
All experiments follow the same training schedule and evaluation protocol described in the main paper unless explicitly noted otherwise.
For the controlled plug-in study, all variants use the same visual inputs, action space, and DUET training objectives, and none of them relies on additional sensors, external maps, external planners, or extra annotation signals (see Table~\ref{tab:plugin_fidelity})~\cite{wu2024cr,fu2019recognition,fu2024understanding}.
For the backbone transfer study, each base model and its S-EGIU-enhanced counterpart are trained and evaluated under the same codebase-specific protocol to ensure fair within-backbone comparison.

The purpose of this supplement is to make the comparison protocol transparent rather than to introduce a separate experimental setting.
Accordingly, the reported results should be interpreted together with the main paper:
the controlled variants localize the contribution of individual mechanism families under a fixed backbone, while the backbone transfer study evaluates whether the proposed state-conditioned instruction updates behave as a transferable mechanism across model architectures.

\endgroup
\end{document}